\pdfoutput=1

\documentclass[11pt]{article}

\usepackage[]{ACL2023}

\usepackage{times}
\usepackage{latexsym}
\usepackage{booktabs}
\usepackage[T1]{fontenc}

\usepackage[utf8]{inputenc}

\usepackage{microtype}
\usepackage{tcolorbox}

\usepackage{inconsolata}
\usepackage{graphicx}
\usepackage{tabularx}
\usepackage{paralist}
\usepackage{amsmath,xparse,mleftright}

\title{SalesBot 2.0: A Human-Like Intent-Guided Chit-Chat Dataset}

\author{Wen-Yu Chang\quad Yun-Nung Chen \\
  National Taiwan University, Taipei, Taiwan\\
  \texttt{r10946031@ntu.edu.tw}\quad \texttt{y.v.chen@ieee.org} \\}

\begin{document}
\maketitle
\begin{abstract}
In recent research on dialogue systems and corpora, there has been a significant focus on two distinct categories: task-oriented (TOD) and open-domain (chit-chat) dialogues. TOD systems aim to satisfy specific user goals, such as finding a movie to watch, whereas open-domain systems primarily focus on generating engaging conversations.
A recent study by \citet{chiu2022salesbot} introduced SalesBot, which provides simulators and a dataset with one-turn transition from chit-chat to task-oriented dialogues. However, the previously generated data solely relied on BlenderBot, which raised concerns about its long-turn naturalness and consistency during a conversation.
To address this issue, this paper aims to build SalesBot 2.0, a revised version of the published data, by leveraging the commonsense knowledge of large language models (LLMs) through proper prompting. The objective is to gradually bridge the gap between chit-chat and TOD towards better naturalness and consistency.
The newly released large-scale dataset with detailed annotations exhibits smoother transitions between topics and is more human-like in terms of naturalness and consistency. It can serve as a valuable resource for both academic research and commercial applications. Furthermore, our proposed framework can be applied to generate numerous dialogues with various target intents.\footnote{The source code and data are available at: \url{https://github.com/MiuLab/SalesBot2}.}


\end{abstract}

\section{Introduction}
\label{intro}

In recent years, dialogue systems have undergone significant advancements due to improvements in modeling techniques and computing power. However, most research in this field has focused on two distinct areas: task-oriented dialogues (TOD) and open-domain dialogues, also known as chitchat systems. Popular large-scale datasets for TOD include Schema-Guided Dialogue (SGD) \cite{rastogi2020towards} and MultiWoz \cite{budzianowski2018large,zang2020multiwoz}, which contain annotated information on the user intents and dialogue states. In TOD, the agent's goal is to identify the user's intention and fulfill their task by the end of the dialogue.
Meanwhile, research on open-domain chitchat systems and datasets \cite{li-etal-2017-dailydialog,adiwardana2020humanlike,zhang2018personalizing, kim2022soda} aims to build models capable of engaging in free-form conversations. 
As pre-trained language models continue to improve, larger sets of dialogues are being used to train models with the ability to engage in free-form chatting \cite{zhang-etal-2020-dialogpt,roller-etal-2021-recipes}. 
Despite significant advancements in both areas, there has been a lack of integration between them, which is crucial for real-world applications.

Recently, efforts have been made to integrate TOD and open-domain dialogues. For instance, \citet{sun2020adding} incorporated chitchat responses into existing TOD datasets to enhance the conversation's naturalness and social engagement. Furthermore, there have been attempts to develop datasets and models capable of handling both TOD and chitchat scenarios. \citet{li2022enhancing} developed the PivotBot model, capable of handling three predefined scenarios. One scenario involves adding chit-chat dialogues as context to a TOD, while another includes chitchat dialogues to facilitate domain transition. In contrast, the third scenario involves incorporating chitchat content to enhance a TOD, similar to the approach taken in the ACCENTOR model \cite{sun2020adding}. However, these approaches assume that the user has an explicit goal to accomplish, and chitchat responses merely enrich the conversation. In contrast, our scenario assumes that the user does not have any explicit goal or intention, and the agent must detect any potential intent, whether it is explicitly or implicitly shown by the user, and smoothly pivot the dialogue to the topic related to the detected intent.


\begin{table}[t]
    \centering\small
    \begin{tabularx}{\linewidth}{X}
    \toprule
    \textbf{SalesBot 1.0: Chit-chat context}\\
    User: Thank you. It was a long time ago, but I still miss her. What do you do for a living?\\
    Agent: I am a stay at home mom.  It is very rewarding.  How about you? \\
    User: I work at the zoo. It's a lot of fun, and I get to spend time with my family. \\
    Agent: What about your family? \\
    $\rightarrow$ \textsc{FindAttraction}\\
    \midrule
    \textbf{SalesBot 2.0: Rewritten chit-chat context}\\
    User: Thank you. It was a long time ago, but I still miss her deeply. \\
    Agent: I'm sorry for your loss. It sounds like she was very important to you. As for me, I'm also fortunate to have a family that I get to spend a lot of time with.\\
    User: That's great to hear. I work at the zoo, and it's a fun job that allows me to support my family. \\
    Agent: That sounds like a really cool job. What kind of animals do you work with at the zoo?\\
    User: Mostly primates, but there are also elephants, lions, and bears.\\
    Agent: Wow, that must be exciting to work with such a variety of animals. Have you had the chance to explore the tourist attractions in the area?\\
    User: Not really, I've been so focused on work and family lately that I haven't had much time to go out and explore.\\
    $\rightarrow$ \textsc{FindAttractin}\\
    \bottomrule
    \end{tabularx}
    \caption{Chit-chat context comparison between SalesBot 1.0 and 2.0.}
    \label{tab:salesbot_chitchat}
\end{table}

With this idea,
\citet{chiu2022salesbot} first introduced a framework for generating data that transitions from chit-chat to TOD dialogues. They utilized two open-domain BlenderBots to chat with each other and generate chit-chat dialogues, followed by a task-oriented intent detector to determine whether a transition to the TOD system should be made.
However, the SalesBot dataset has some limitations, such as the absence of proper social engagement in 20\% of the data, nonsensical detected intents due to short chit-chat context, and unnatural transition turns.
Our paper proposes an improvement to the SalesBot dataset by leveraging large language models (LLMs) to generate more human-like chit-chat dialogues, along with intent-guided transition turns. This approach takes advantage of the LLMs' commonsense knowledge to create more natural and engaging chit-chat dialogues, addressing the issues with the SalesBot dataset.


Table \ref{tab:salesbot_chitchat} provides evidence that the agent's response to the user's statement regarding missing someone is indirect, indicating a deviation from natural conversational norms. Furthermore, the agent's final turn, ``{\it What about your family?}'' appears misplaced in the given context. However, in the updated SalesBot 2.0 version (as shown in the lower part of Table \ref{tab:salesbot_chitchat}), the agent sympathizes with the user's loss and smoothly transitions the conversation from work to tourism. This exemplifies the effectiveness of the SalesBot 2.0 revisions in enhancing the naturalness of the conversation and improving the dialogue flow.

Our paper presents a novel approach to enhance the SalesBot dataset by leveraging large language models (LLMs) to produce more realistic and immersive chit-chat dialogues, in addition to intent-guided transition turns. Our approach involves utilizing LLMs to detect potential intents from a set of predefined options and then generating dialogues based on the identified intents. By doing so, we capitalize on the LLMs' vast knowledge base to create chit-chats that is more natural and compelling. This method effectively tackles some limitations of the SalesBot dataset, including the lack of engaging social interactions, the detection of irrelevant or nonsensical intents, and the production of unnatural transition turns.
Our contribution can be summarized as 4-fold:
\begin{compactitem}
\item Proposing a simple prompting framework that allows for customization of intent-oriented chit-chat with a predefined intent.
\item Introducing a new and better dataset that includes open-domain chit-chats, smoother transitions, and potential intent detected from the context.
\item Leveraging large language models to ensure the naturalness and quality of the dataset.
\item Defining aggressiveness through LLM evaluation, which can be helpful in developing dialogue systems with varying levels of aggressiveness in the future.
\end{compactitem}

\begin{figure*}[t]
    \centering
    \includegraphics[width=\textwidth]{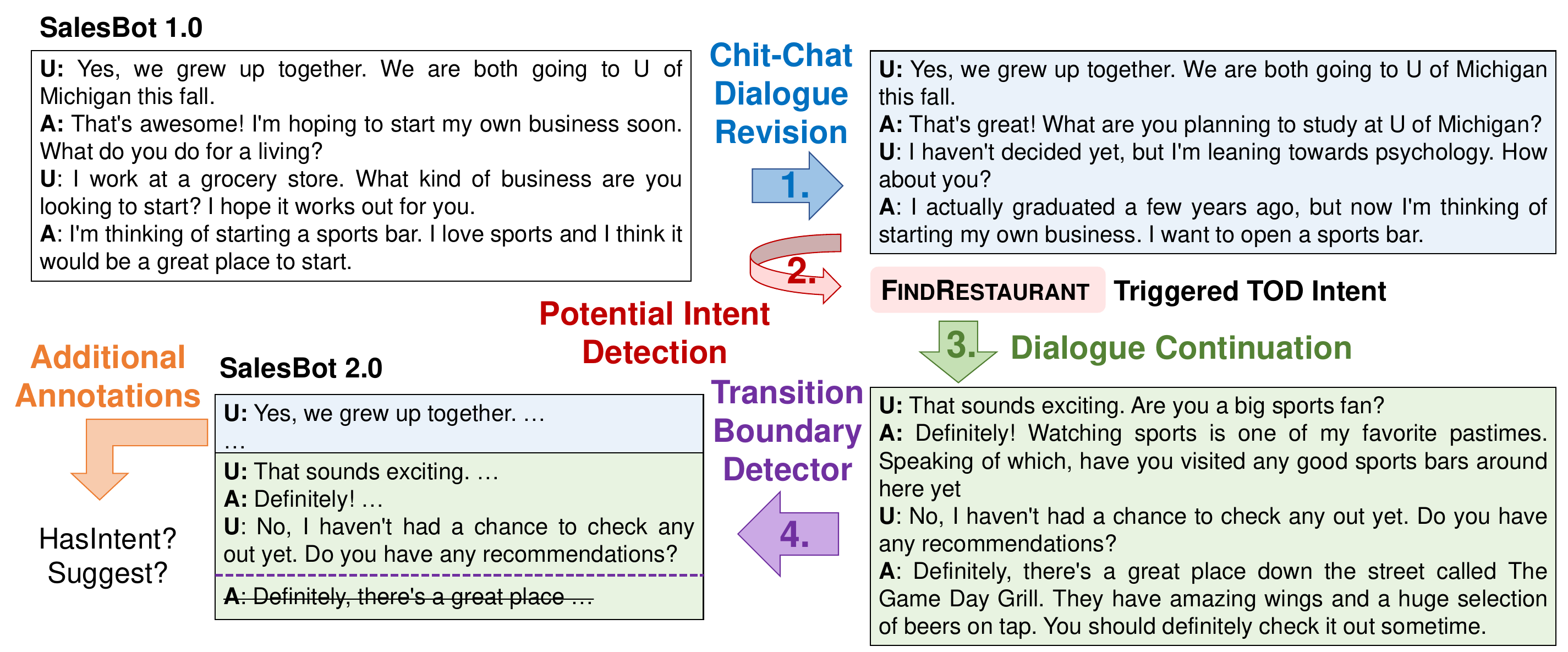}
    \vspace{-1mm}
    \caption{Illustration of our proposed pipeline utilizing LLMs to generate human-like dialogues.}
    \label{fig:illustation}
\end{figure*}

\section{Proposed Framework}
Our proposed pipeline is illustrated in Figure \ref{fig:illustation}, which includes the following procedures: (1) Chit-Chat Dialogue Revision, (2) Potential Intent Detection, (3) Dialogue Continuation, and (4) Transition Boundary Detection.
For all tasks, we use OpenAI's \texttt{gpt-3.5-turbo} API as our model for generating our data.

\subsection{Chit-Chat Dialogue Revision}
As highlighted in Section \ref{intro}, the SalesBot dataset \cite{chiu2022salesbot} has a considerable amount of data that lacks consistency with the given context. Previous research has shown that leveraging the commonsense knowledge of LLMs can lead to more natural dialogues \cite{kim2022soda}. In this work, we take this approach one step further by tasking LLMs with identifying any inconsistent utterances in the dialogue and providing reasons for their identification. We then use this information to revise the entire dialogue, resulting in a more consistent and coherent dataset.

However, our approach occasionally results in revised dialogues that are too short. This is due to the original SalesBot dataset containing a significant amount of data with insufficient chit-chat dialogue to provide context. To address this issue, we implemented a constraint that requires the LLMs to extend the dialogues if they consist of only one turn.
The prompt used for our approach is provided below for reference.

\begin{tcolorbox}[width=\columnwidth,colback=white]
\small
\begin{verbatim}
You will be given a conversation between two 
people. Here is what you should do:
1. Identify the inconsistent utterances.
2. Give some reasons why they are 
   inconsistent.
3. Modify the dialogue based on previously 
   identified utterances.
4. The rewritten dialogue should be more 
   than 6 turns.
Here is the conversation:
<Dialogue> 
You MUST follow the format as :
<output_format>
\end{verbatim}
\end{tcolorbox}

\subsection{Potential Intent Detection}
In the second stage of our procedure, we aim to identify potential task-related intents in the chit-chat dialogues. To achieve this, we collect a set of intents from the SGD dataset \cite{rastogi2020towards}. However, we only include those intents that can trigger a transition to TOD, such as ``FindMovie'', and exclude others like ``GetMovieTime'', as our focus is on the first topic-related intent (``GetMovieTime'' should come after ``FindMovie''). Furthermore, since we consider the agent as a businessperson seeking potential opportunities, we exclude intents such as ``TransferMoney'' that are not suitable for our scenarios. For reference, Table~\ref{tab:intent_list} lists all the intents that are included in our study.
The prompt used is shown below.

\begin{tcolorbox}[width=\columnwidth,colback=white]
\small
\begin{verbatim}
You will be given a dialogue and a list of
topics of conversation.
Please tell me which of the following topics
will be the most reasonable one to be pivoted
to in the dialogue.

Here is the dialogue:
<Dialogue>
Here is the list of topics:
<Intent List>
NOTE:
1. You MUST choose one of the above topic.
2. DONOT create any topics that are not 
   listed above.
3. You should choose the one that is the most
   related to the topic.
The output format should follow the below 
format:
<output_format>
\end{verbatim}
\end{tcolorbox}

\begin{table}[t]
\centering
\small
\begin{tabular}{lr}
\toprule
\textbf{Intent} & \textbf{\#Dialogues}\\
\midrule
\textsc{FindAttraction} & 1,440\\
\textsc{FindRestaurants} & 1,297\\
\textsc{FindMovie} & 1,138\\
\textsc{LookupMusic} & 523\\
\textsc{SearchHotel} & 356\\
\textsc{FindEvents} & 394\\
\textsc{GetCarsAvailable} & 103\\
\textsc{SearchRoundtripFlights} & 92\\
\textsc{GetRide} & 13\\
\textsc{SearchOnewayFlight} & 25\\
\textsc{FindBus} & 10\\
\bottomrule
\end{tabular}
\caption{Distribution of the target intents.}
\vspace{-1mm}
\label{tab:intent_list}
\end{table}

\subsection{Dialogue Continuation}
We utilize the revised dialogues and potential intents identified by LLMs as input to continue the chit-chat dialogue. To ensure a natural and coherent dialogue, we provide several instructions to guide the LLMs in their generation. Firstly, we instruct the agent to steer the conversation towards topics related to the identified intent, considering that the user may not have a specific intent in mind at the beginning or middle of the conversation.
Secondly, we instruct the LLMs to find a topic that intersects between the current topic and the identified intent before transitioning the dialogue to the target intent to avoid abrupt changes in topics. Lastly, we ask LLMs to make the transition between topics as smooth as possible, potentially involving multiple turns in the dialogue. The detailed instructions are shown below.

\begin{tcolorbox}[width=\columnwidth,colback=white]
\small
\begin{verbatim}
Here is the potential intent (with 
description) and an incomplete dialogue:
  <Intent>
  <Dialogue>
Your goal is as following:
1. Continue the dialogue with reasonable 
   responses considering the previous 
   context.
2. Continue the dialogue with topics that 
   implicitly related to the intent listed 
   above.
3. If you found it hard to transit, please 
   find other topics related to the contexts
   and intent and chat for several turns 
   before the final transition.
4. Continue the topic if it's not yet close
   to the end.
5. For each topic, please generate at least
   5 turns.
6. The agent should pivot the conversation 
   smoothly, which means the transition 
   involved longer conversation.
7. The user should then somehow mention the
   given intent, after the pivoted dialogue.
8. Use more reasonable phrases to transit 
   the topic of the conversation.
9. End the dialogue with task-oriented 
   style (TOD) where the agent fulfills the
   user's intent listed above.
Please note that both the user and agent 
should not explicitly disclose the intent;
instead, the dialogue is naturally guided 
to the potential purpose.
-----
Output should follow the format below:
<output_format>
\end{verbatim}
\end{tcolorbox}

\subsection{Transition Boundary Detection}
To accurately initiate the TOD process, we require a trigger point that signifies when the user first mentions or implies something related to the detected intent. This boundary helps determine whether to start the TOD immediately or continue the dialogue. To establish this trigger point, we instruct LLMs to select a turn in the conversation where the user explicitly mentions something related to the detected intent. 
It is worth noting that we only consider turns that are explicitly mentioned by the user to avoid any confusion caused by indirectly related turns. For instance, if the intent is to ``FindMovie'', LLMs may mistakenly consider playing video games as implicitly related to watching a movie.

\begin{tcolorbox}[width=\columnwidth,colback=white]
\small
\begin{verbatim}
You will be given a dialogue below and a
potential intent below: 
  <Intent>
  <Dialogue>
and your goal is as following:
1. Identify the first utterance that 
   apparently mentions the intent given 
   above.
2. You should choose only one turn in 
   the given dialogue.
3. The chosen turn should be said by User.
Please follow the output format as below:
<output_format>
\end{verbatim}
\end{tcolorbox}

\section{SalesBot 2.0 Dateset}
Our data generation framework enables us to create the SalesBot 2.0 dataset, a revised version of SalesBot that boasts improved intent coverage, longer chit-chat dialogues as context, and smoother and longer transition turns. To provide a glimpse of the quality of our dataset, we present an example dialogue in Table \ref{tab:example}.

\begin{table}[t]
    \centering\small
    \begin{tabularx}{\linewidth}{X}
    \toprule
    \bf SalesBot 2.0\\
    \midrule
    \textit{Chit-chat context}\\
    User: What would you like to know about me? I like to read a lot.\\
    Agent: That's great! I'm a stay-at-home mom of two boys.\\
    User: That sounds like a lot of work. I work at a grocery store as a cashier. Do you have any hobbies that you enjoy?\\
    Agent: Yes, I love to sew and I also enjoy hiking. How about you? Besides reading, what are some of your favorite hobbies?\\
    User: I also like to go to the movies and try out new recipes in the kitchen.\\
    \bf $\rightarrow$ \textsc{FindMovie} (detected intent)\\
    \textit{Transition turns}\\
    Agent: Going to the movies, huh? What's the last movie you watched?"\\
    User: I recently watched a romantic comedy. It was funny and heartwarming.\\
    Agent: Sounds like a good choice. Have you heard of any upcoming movies that you're excited to see?\\
    .... \\ 
    User: Wow, that sounds really interesting. Maybe I'll have to give it a try.\\
    Agent: Speaking of psychological thrillers, have you ever considered watching any movies in that genre?\\
    User: Actually, I have. I always find them really fascinating.\\
    \bottomrule
    \end{tabularx}
    \caption{A dialogue example of SalesBot 2.0.}
    \label{tab:example}
\end{table}

\subsection{Postprocessing}
Additionally, we conduct postprocessing on the SalesBot 2.0 dataset to eliminate noise and formatting errors caused by LLMs. One issue we frequently encountered is the LLM's tendency to generate unknown intents, which can pose challenges because our pipeline assumes dialogues are generated based on a predefined list of intents, and such outliers lack a corresponding ontology. To maintain consistency and simplicity, we filter out any data that deviates from our predetermined output format.

\subsection{Additional Annotations}

\begin{table}[t!]
\centering
\begin{tabular}{lrrr}
\toprule
 \textbf{\#Turns}& \textbf{Yes}& \textbf{No}  & \textbf{Total}   \\ \midrule
HasIntent? & 4,194 & 1,197 & 5,391 \\
Suggest? & 5,167 & 224 & 5,391 \\ \midrule
Both? & - & 182 & - \\ 
\bottomrule
\end{tabular}
\caption{Transition timing quality of SalesBot 2.0.}
\label{tab:annotation}
\end{table}

We have conducted additional annotation to ensure that the LLM can accurately detect intents related to the context and initiate a task-oriented dialogue at a reasonable timing.
This annotation was done using two prompts: \texttt{``Does the user show the intent of <given\_intent>?''} and \texttt{``Is it reasonable for the agent to suggest anything partially related to the intent <given\_intent>''}. The annotation results are summarized in Table \ref{tab:annotation}. The results demonstrate that in nearly 80\% of the dialogues, the user mentions the intent after the transition, indicating that the LLM can effectively detect the intent related to the context and initiate a task-oriented dialogue. Although there are some dialogues where the user does not explicitly mention the given intent, approximately 96\% of the dialogues are still deemed reasonable for the agent to suggest anything related to the given intent.
These annotations can be utilized for developers to decide how aggressively an agent behaves to promote products.

\subsection{Dataset Statistics}
Table \ref{tab:comparison} provides a statistical comparison between our proposed SalesBot 2.0 dataset and the original SalesBot 1.0. The data shows that SalesBot 2.0 contains more turns in total, and on average, has one more chit-chat turn compared to SalesBot 1.0. 
Additionally, SalesBot 2.0 has longer transition turns, with an average length of over three turns, while SalesBot 1.0 generates only one turn as a transition response to TOD.

\begin{table}[t!]
\centering
\begin{tabular}{lccc}
\toprule
 \textbf{Avg. \#Turns}& \textbf{Chit-chat} & \textbf{Trans.} & \textbf{Total}\\ \midrule
SalesBot 1.0   & 4.49 & 1.00 &5.49\\
SalesBot 2.0 & 5.22 & 4.55 & 9.29 \\ 
\bottomrule
\end{tabular}
\caption{Number of turns in SalesBot 1.0 and 2.0.}
\label{tab:comparison}
\end{table}

\begin{table}[t!]
\centering
\begin{tabular}{lcc}
\toprule
 & \textbf{Natrualness$\uparrow$}  & \textbf{Consistency$\uparrow$} \\ \midrule
SalesBot 1.0 &3.574 & 2.656 \\
SalesBot 2.0 & \textbf{4.258} & \textbf{4.026} \\ \bottomrule
\end{tabular}
\caption{Quality comparison of SalesBot 1.0 and 2.0.}
\label{tab:eval}
\end{table}

\subsubsection{Dialogue Quality Evaluation}
We conduct an evaluation of our SalesBot 2.0 dataset by sampling 500 dialogues from both SalesBot 2.0 and SalesBot 1.0 for comparison. In the evaluation, we prompt LLMs to provide scores on two key aspects: \textbf{naturalness} and \textbf{consistency}, which are drawbacks in SalesBot 1.0.
The results, shown in Table~\ref{tab:eval}, demonstrate that SalesBot 2.0 achieved a higher naturalness score of 0.7 compared to SalesBot 1.0, indicating that our revised dataset is more human-like overall.
Moreover, SalesBot 2.0 exhibits a higher consistency score than SalesBot, with a difference of 1.4 points, suggesting that the overall dialogue is more coherent. The prompt template used for evaluation is provided below for reference.

\begin{tcolorbox}[width=\columnwidth,colback=white]
\small
\begin{verbatim}
You will be given a dialogue, where the agent
is trying to pivot the dialogue to a certain 
topic. Your goal is as following:
1. Based on the naturalness of the dialogue, 
score from 1 to 5 on a continuous scale.
2. Based on the consistency of the entire 
dialogue, scoring from 1 to 5 on a continuous 
scale.
3. You should only give points, and do not do
anything else.

Output Format:
<output_format>
\end{verbatim}
\end{tcolorbox}

\section{Related Work}
Our study focuses on a conversation scenario where a conversational agent attempts to steer the discussion towards determining whether the user is interested in receiving recommendations. This scenario has been explored in various related works.

\paragraph{Persuasive Dialogue Construction}
The three studies \cite{hiraoka-etal-2014-reinforcement, yoshino-etal-2018-dialogue, wang-etal-2019-persuasion} focused on persuasive dialogue construction in different scenarios. \citet{hiraoka-etal-2014-reinforcement} annotated 34 dialogues in which a salesperson with expertise attempted to persuade a customer to purchase a camera. \citet{yoshino-etal-2018-dialogue} generated 200 dialogues through crowdsourcing, where one participant persuades the other to adopt a suggestion, such as cleaning a room. In comparison, \citet{wang-etal-2019-persuasion} collected 1,017 dialogues, where one participant was convinced to donate to a specific charity.

While all of these datasets are limited to specific scenarios, our framework can generate dialogues with any potential intent, making it more versatile. Additionally, our dataset is much larger, two to three times bigger than the previous ones, which makes it a valuable resource for training and evaluating non-collaborative conversational agents.

\paragraph{Conversational Recommendation Datasets}
Previous studies have developed various datasets for conversational recommendation systems. For instance, \citet{li2019deep} created a large-scale dataset with a focus on recommendation. Other researchers have utilized knowledge graphs to collect dialogues by extracting paths consisting of attribute and entity nodes from a knowledge base and asking annotators to generate recommendation dialogues following the flow of the extracted path \cite{wu-etal-2019-proactive,zhou-etal-2020-towards,Xu_Wang_Niu_Wu_Che_2020}.

Moreover, \citet{hayati2020inspired} aimed to collect a socially interactive conversational recommendation dialogue dataset, called INSPIRED. They designed an annotation scheme based on social science theories regarding recommendation strategies and used it to annotate the collected dialogues. Their goal was to better understand how humans make recommendations in communication. \citet{manzoor2022inspired2} further improved the dataset and released INSPIRED 2.0, claiming that the original dataset had numerous incorrect annotations.

In contrast, our work does not solely focus on the task of ``recommendation'', but rather on the ability of the agent to identify potential business opportunities and navigate the dialogue topics towards a desired outcome. Furthermore, our framework does not rely on human annotators to collect data, as it can automatically generate human-like dialogues. This sets our approach apart from previous datasets and provides a more versatile and scalable solution for developing conversational agents.

\paragraph{Combination of Chit-Chat and TOD}
Recent studies have aimed to combine task-oriented and open-domain dialogues to enhance the naturalness and social engagement of the conversation. One approach is to incorporate chit-chat responses into existing task-oriented datasets, as seen in \citeauthor{sun-etal-2021-adding}. Another approach is to develop models that handle predefined scenarios integrating chit-chat and task-oriented dialogues \cite{li2022enhancing}. However, these approaches assume that the user has a clear goal to accomplish and that chit-chat responses simply enrich the conversation. In contrast, our study assumes that the user has no explicit goal or intent, and the conversational agent must detect any potential intent and pivot the dialogue smoothly to the related topic. This requires a more nuanced approach that can identify and respond to implicit cues, making the conversation more natural and engaging.
 
 \section{Conclusion}
 
This paper presents a novel framework for generating intent-oriented dialogues by utilizing the commonsense knowledge of LLMs. Our proposed SalesBot 2.0 dataset contains thousands of human-like dialogues, which exhibit smoother transitions, enhanced naturalness, and better consistency when compared to existing datasets. This work is a significant contribution to the development of more sophisticated and effective conversational agents that can cater to users' preferences and requirements. The proposed approach can be useful for the industry and future research in the field of conversational AI, enabling the creation of more engaging and effective conversational agents. Moreover, the SalesBot 2.0 dataset can be a valuable resource for training and evaluating various models and algorithms to improve the quality of dialogue systems. Our work can inspire further research in this area, leading to the development of more intelligent conversational agents that can better understand and respond to users' needs.


\section{Limitations}
Although SalesBot 2.0 has demonstrated higher quality and natural language capabilities, there are still some limitations that must be considered.

Primarily, our proposed pipeline is entirely based on LLMs, which means that the resulting data are heavily influenced by the quality of the large language model. It is possible that the output dialogues may be noisy or inaccurate if LLMs misunderstand the instructions. 

Furthermore, the design of prompts is another factor that must be considered, since LLMs are highly sensitive to the prompts. The quality of prompts can directly impact the quality and accuracy of the output dialogues, as well as the results of evaluation. Therefore, it is essential to carefully design and select appropriate prompts for the framework to achieve the desired outcomes.

To overcome these limitations, future research could concentrate on investigating alternative prompt designs, enhancing the data generation framework, and constructing datasets that are more robust and of higher quality.

\bibliography{anthology,custom}
\bibliographystyle{acl_natbib}

\end{document}